\documentclass[11pt,a4paper]{article}
\usepackage[hyperref]{naaclhlt2019}
\usepackage{times}
\usepackage{latexsym}
\usepackage{subcaption}
\usepackage{graphicx}

\usepackage{url}

\aclfinalcopy % Uncomment this line for the final submission
\title{Lipstick on a Pig: \\Debiasing Methods Cover up Systematic Gender Biases\\ in Word Embeddings But do not Remove Them}

\author{Hila Gonen$^1$ \and Yoav Goldberg$^{1,2}$ \\
	$^1$Department of Computer Science,  Bar-Ilan University \\
	$^2$Allen Institute for Artificial Intelligence \\
	{\tt \{hilagnn,yoav.goldberg\}@gmail.com} \\
}

\date{}

\begin{document}
	\maketitle
	\begin{abstract}
		Word embeddings are widely used in NLP for a vast range of tasks. It was shown that word embeddings derived from text corpora reflect gender biases in society. This phenomenon is pervasive and consistent across different word embedding models, causing serious concern. Several recent works tackle this problem, and propose methods for significantly reducing this gender bias in word embeddings, demonstrating convincing results. 
		However, we argue that this removal is superficial. While the bias is indeed substantially reduced according to the provided bias definition, the actual effect is mostly hiding the bias, not removing it. The gender bias information is still reflected in the distances between ``gender-neutralized'' words in the debiased embeddings, and can be recovered from them. We present a series of experiments to support this claim, for two debiasing methods. We conclude that existing bias removal techniques are insufficient, and should not be trusted for providing gender-neutral modeling.
	\end{abstract}
	
	\section{Introduction}
	\label{Introduction}

	Word embeddings have become an important component in many NLP models and are widely used for a vast range of downstream tasks. However, these word representations have been proven to reflect social biases (e.g. race and gender) that naturally occur in the data used to train them \cite{CBN17,GSJ18}. 
	
	In this paper we focus on gender bias. Gender bias was demonstrated to be consistent and pervasive across different word embeddings. Bolukbasi et al. \citeyearpar{BCZ16} show that using word embeddings for simple analogies surfaces many gender stereotypes. For example, the word embedding they use (word2vec  embedding trained on the Google News dataset\footnote{\scriptsize{\url{https://code.google.com/archive/p/word2vec/}}} \cite{MCC13}) answer the analogy ``man is to computer programmer as woman is to x" with ``x = homemaker".	Caliskan et al. \citeyearpar{CBN17} further demonstrate association between female/male names and groups of words stereotypically assigned to females/males (e.g. arts vs. science). In addition, they demonstrate that word embeddings reflect actual gender gaps in reality by showing the correlation between the gender association of occupation words and labor-force participation data.
	
	Recently, some work has been done to reduce the gender bias in word embeddings, both as a post-processing step \cite{BCZ16} and as part of the training procedure \cite{ZZL18}. Both works substantially reduce the bias with respect to the same definition: the projection on the gender direction (i.e. $\overrightarrow{he} - \overrightarrow{she}$), introduced in the former. They also show that performance on word similarity tasks is not hurt.
	
	We argue that current debiasing methods, which lean on the above definition for gender bias and directly target it, are mostly hiding the bias rather than removing it. We show that even when drastically reducing the gender bias according to this definition, it is still reflected in the geometry of the representation of ``gender-neutral" words, and a lot of the bias information can be recovered.\footnote{The code for our experiments is available at \url{https://github.com/gonenhila/gender_bias_lipstick}.}
	
	\section{Gender Bias in Word Embeddings}
	\label{background}
	
	In what follows we refer to words and their vectors interchangeably.
	
	\paragraph{Definition and Existing Debiasing Methods}
	
	Bolukbasi et al. \citeyearpar{BCZ16} define the gender bias of a word $w$ by its projection on the ``gender direction": $\overrightarrow{w} \cdot (\overrightarrow{he} - \overrightarrow{she})$, assuming all vectors are normalized. The larger a word's projection is on $\overrightarrow{he} - \overrightarrow{she}$, the more biased it is. They also quantify the bias in word embeddings using this definition and show it aligns well with social stereotypes. 
	
	Both Bolukbasi et al. \citeyearpar{BCZ16} and Zhao et al. \citeyearpar{ZZL18} propose methods for debiasing word embeddings, substantially reducing the bias according to the suggested definition.\footnote{Another work in this spirit is that of Zhang et al. \citeyearpar{ZLM18}, which uses an adversarial network to debias word embeddings. There, the authors rely on the same definition of gender bias that considers the projection on the gender direction. We expect similar results for this method as well, however, we did not verify that.} 
	
	In a seminal work, Bolukbasi et al. \citeyearpar{BCZ16} use a post-processing debiasing method. Given a word embedding matrix, they make changes to the word vectors in order to reduce the gender bias as much as possible for all words that are not inherently gendered (e.g. mother, brother, queen). They do that by zeroing the gender projection of each word on a predefined gender direction.\footnote{The gender direction is chosen to be the top principal component (PC) of ten gender pair difference vectors.} In addition, they also take dozens of inherently gendered word pairs and explicitly make sure that all neutral words (those that are not predefined as inherently gendered) are equally close to each of the two words. This extensive, thoughtful, rigorous and well executed work surfaced the problem of bias in embeddings to the ML and NLP communities, defined the concept of debiasing word embeddings, and established the defacto metric of measuring this bias (the gender direction). It also provides a perfect solution to the problem of removing the gender direction from non-gendered words. However, as we show in this work, while the gender-direction is a great indicator of bias, it is only an indicator and not the complete manifestation of this bias.
	
	Zhao et al. \citeyearpar{ZZL18} take a different approach and suggest to train debiased word embeddings from scratch. Instead of debiasing existing word vectors, they alter the loss of the GloVe model \cite{PSM14}, aiming to concentrate most of the gender information in the last coordinate of each vector. This way, one can later use the word representations excluding the gender coordinate. They do that by using two groups of male/female seed words, and encouraging words that belong to different groups to differ in their last coordinate. In addition, they encourage the representation of neutral-gender words (excluding the last coordinate) to be orthogonal to the gender direction.\footnote{The gender direction is estimated during training by averaging the differences between female words and their male counterparts in a predefined set.}
	This work did a step forward by trying to remove the bias during training rather than in post-processing, which we believe to be the right approach. Unfortunately, it relies on the same definition that we show is insufficient.
	
	\noindent\textbf{These works implicitly define what is good gender debiasing:} according to \citet{BCZ16}, there is no
	gender bias if each non-explicitly gendered word in the vocabulary is in equal distance to both elements of all explicitly gendered pairs. In other words, if one cannot determine the gender association of a word by looking at its projection on any gendered pair. In \citet{ZZL18} the definition is similar, but restricted to projections on the gender-direction.

	\paragraph{Remaining bias after using debiasing methods}
	
	Both works provide very compelling results as evidence of reducing the bias without hurting the performance of the embeddings for standard tasks. 
	
	However, both methods and their results rely on the specific bias definition. We claim that the bias is much more profound and systematic, and that simply reducing the projection of words on a gender direction is insufficient: it merely hides the bias, which is still reflected in similarities between ``gender-neutral" words (i.e., words such as ``math'' or ``delicate'' are in principle gender-neutral, but in practice have strong stereotypical gender associations, which reflect on, and are reflected by, neighbouring words). 
	
	Our key observation is that, almost by definition, most word pairs maintain their previous similarity, despite their change in relation to the gender direction. The implication of this is that most words that had a specific bias before are still grouped together, and apart from changes with respect to specific gendered words, the word embeddings' spatial geometry stays largely the same.\footnote{We note that in the extended arxiv version, \citet{BCZSK16} do mention this phenomenon and refer to it as ``indirect bias". However, they do not quantify its extensiveness before and after debiasing, treat it mostly as a nuance, and do not provide any methods to deal with it.}	
	In what follows, we provide a series of experiments that demonstrate the remaining bias in the debiased embeddings.
	
	\section{Experimental Setup}
	\label{experiments}
	
	We refer to the word embeddings of the previous works as \textsc{Hard-Debiased} \cite{BCZ16} and \textsc{GN-GloVe} (gender-neutral GloVe) \cite{ZZL18}. For each debiased word embedding we quantify the hidden bias with respect to the biased version. For \textsc{Hard-Debiased} we compare to the embeddings before applying the debiasing procedure. For \textsc{GN-Glove} we compare to embedding trained with standard GloVe on the same corpus.\footnote{We use the embeddings provided by Bolukbasi et al. \citeyearpar{BCZ16} in \url{https://github.com/tolga-b/debiaswe} and by Zhao et al. \citeyearpar{ZZL18} in \url{https://github.com/uclanlp/gn_glove}.}
	
	Unless otherwise specified, we follow \citet{BCZ16} and use a reduced version of the vocabulary for both word embeddings: we take the most frequent 50,000 words and phrases and remove words with upper-case letters, digits, or punctuation, and words longer than 20 characters. In addition, to avoid quantifying the bias of words that are inherently gendered (e.g. mother, father, queen), we remove from each vocabulary the respective set of gendered words as pre-defined in each work.\footnote{For \textsc{Hard-Debiased} we use first three lists from: \url{https://github.com/tolga-b/debiaswe/tree/master/data} and for \textsc{GN-GloVe} we use the two lists from: \url{https://github.com/uclanlp/gn_glove/tree/master/wordlist}} This yeilds a vocabulary of 26,189 words for \textsc{Hard-Debiased} and of 47,698 words for \textsc{GN-GloVe}.
	
	As explained in Section~\ref{background} and according to the definition in previous works, we compute the bias of a word by taking its projection on the gender direction: $\overrightarrow{he} - \overrightarrow{she}$.
	
	In order to quantify the association between sets of words, we follow Caliskan et al. \citeyearpar{CBN17} and use their Word Embedding Association Test (WEAT): consider two sets of target words (e.g., male and female professions) and two sets of attribute words (e.g., male and female names). A permutation test estimates the probability that a random permutation of the target words would produce equal or greater similarities to the attribute sets. 		
	
	\section{Experiments and Results}
	
\begin{figure}
		\centering
		
		\begin{subfigure}[b]{1\linewidth}
			\includegraphics[width=1\linewidth]{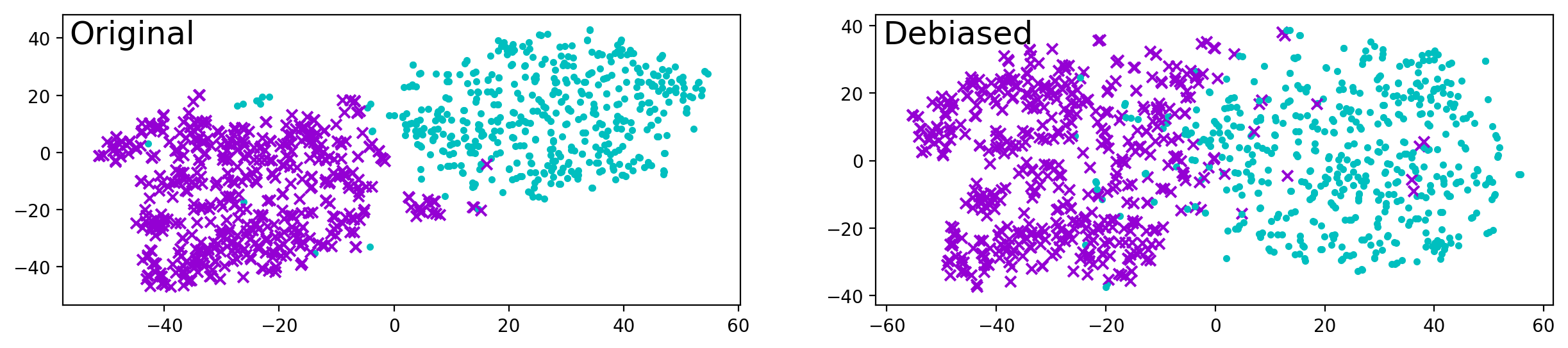}
			\caption{Clustering for \textsc{Hard-Debiased} embedding, before (left hand-side) and after (right hand-side) debiasing.}
		\end{subfigure}
		
		\begin{subfigure}[b]{1\linewidth}
			\includegraphics[width=1\linewidth]{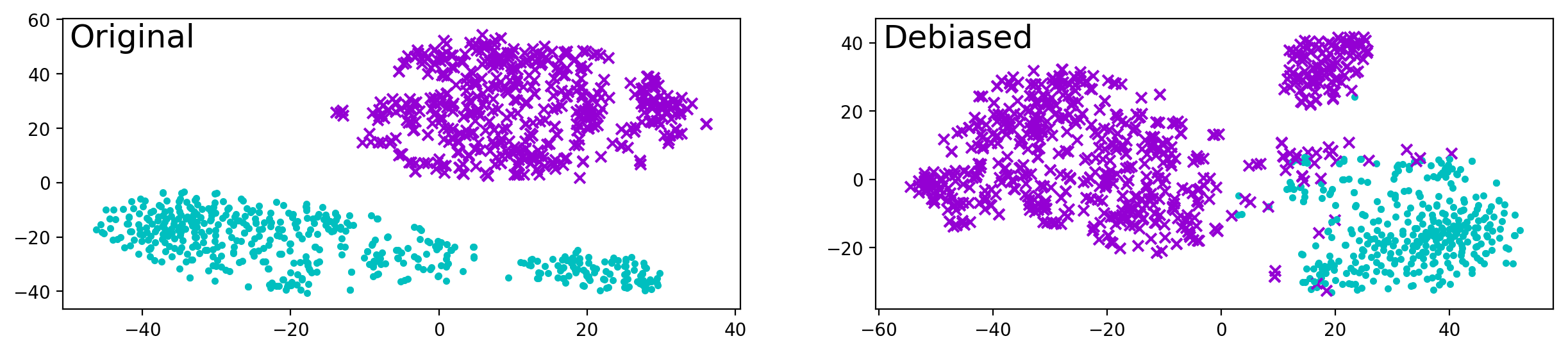}
			\caption{Clustering for \textsc{GN-GloVe} embedding, before (left hand-side) and after (right hand-side) debiasing.}
			
		\end{subfigure}
		
		\caption{Clustering the 1,000 most biased words, before and after debiasing, for both models.}
		\label{cluster}
	
	\end{figure}

	\paragraph{Male- and female-biased words cluster together}
	We take the most biased words in the vocabulary according to the original bias (500 male-biased and 500 female-biased\footnote{highest on the two lists for \textsc{Hard-Debiased} are 'petite', 'mums', 'bra', 'breastfeeding' and 'sassy' for female and 'rookie', 'burly', 'hero', 'training\_camp' and 'journeyman' for male. Lowest on the two lists are 'watchdogs', 'watercolors', 'sew', 'burqa', 'diets' for female and 'teammates', 'playable', 'grinning', 'knee\_surgery', 'impersonation' for male.}), and cluster them into two clusters using k-means. For the \textsc{Hard-Debiased} embedding, the clusters align with gender with an accuracy of 92.5\% (according to the original bias of each word), compared to an accuracy of 99.9\% with the original biased version. For the \textsc{GN-GloVe} embedding, we get an accuracy of 85.6\%, compared to an accuracy of 100\% with the biased version. These results suggest that indeed much of the bias information is still embedded in the representation after debiasing. Figure~\ref{cluster} shows the tSNE \cite{MH08} projection of the vectors before and after debiasing, for both models.

	\begin{figure*}[ht]
	\centering
	
	\begin{subfigure}{0.45\linewidth}
		\includegraphics[width=\linewidth]{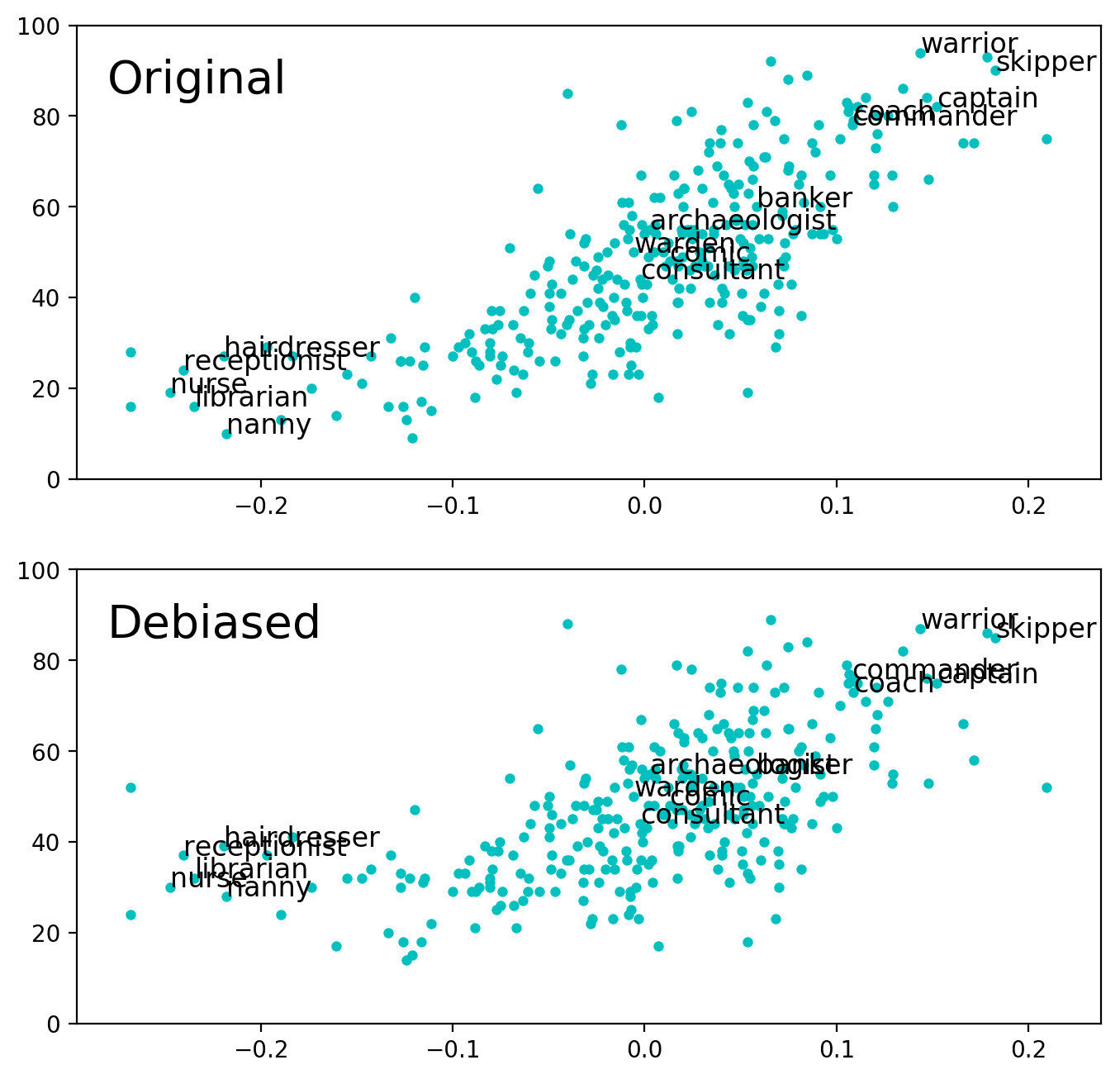}
		\caption{The plots for \textsc{Hard-Debiased} embedding, before (top) and after (bottom) debiasing.}
	\end{subfigure} \hspace{0.05\textwidth}
	\begin{subfigure}{0.45\linewidth}
		\includegraphics[width=\linewidth]{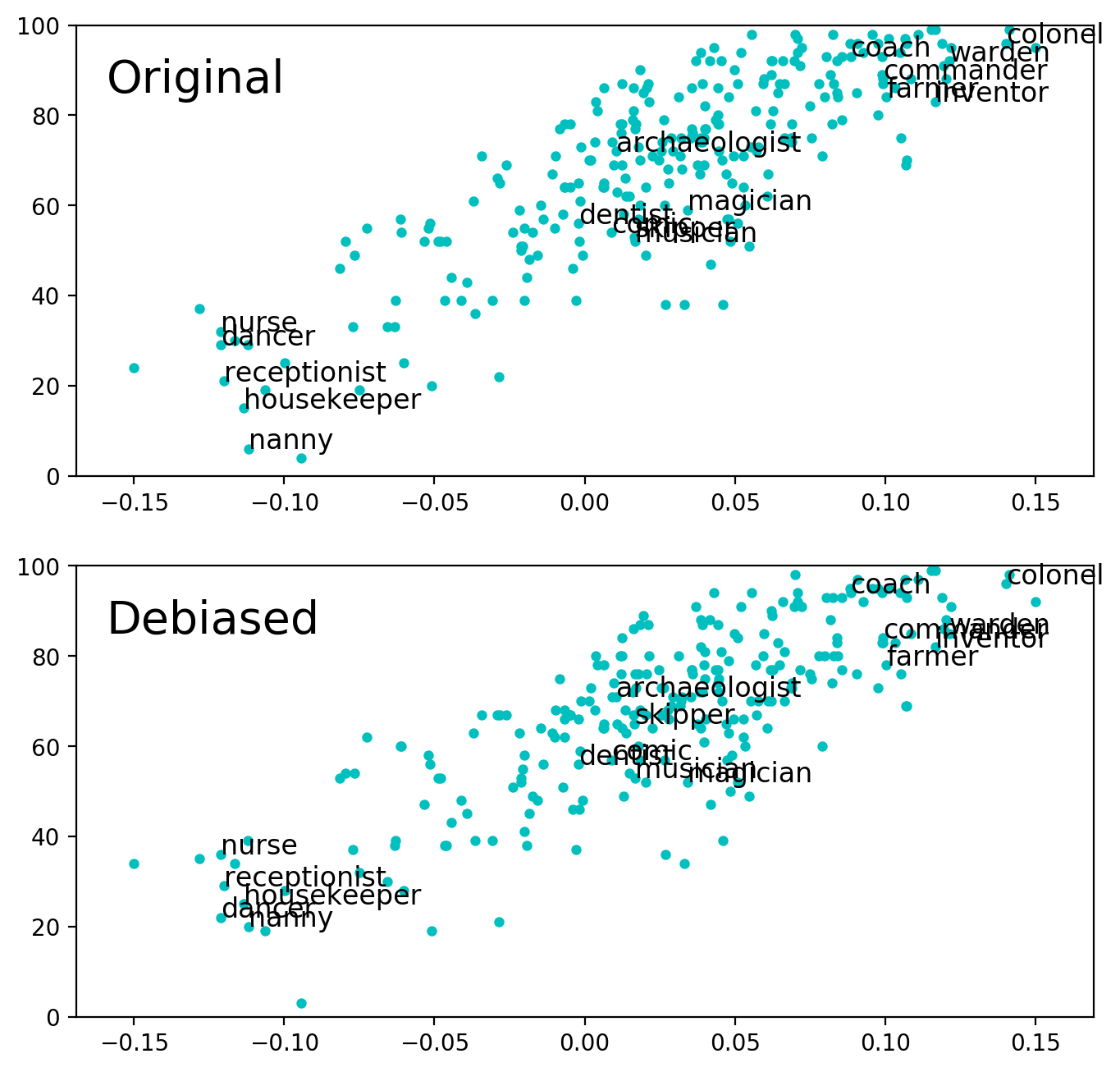}
		\caption{The plots for \textsc{GN-GloVe} embedding, before (top) and after (bottom) debiasing.}
	\end{subfigure}
	
	\caption{The number of male neighbors for each profession as a function of its original bias, before and after debiasing. We show only a limited number of professions on the plot to make it readable.}
	\label{prof_graphs}
	
	\end{figure*}

	\paragraph{Bias-by-projection correlates to bias-by-neighbours}
	This clustering of gendered words indicates that while we cannot directly ``observe" the bias (i.e. the word ``nurse" will no longer be closer to explicitly marked feminine words) the bias is still manifested by the word being close to \emph{socially-marked} feminine words, for example ``nurse" being close to ``receptionist", ``caregiver" and ``teacher". This suggests a new mechanism for measuring bias: the percentage of male/female socially-biased words among the k nearest neighbors of the target word.\footnote{While the social bias associated with a word cannot be observed directly in the new embeddings, we can approximate it using the gender-direction in non-debiased embeddings.}
	
	We measure the correlation of this new bias measure with the original bias measure. For the \textsc{Hard-Debiased} embedding we get a Pearson correlation of 0.686 (compared to a correlation of 0.741 when checking neighbors according to the biased version). For the \textsc{GN-GloVe} embedding we get a Pearson correlation of 0.736 (compared to 0.773). All these correlations are statistically significant with p-values of 0. 
	
	\paragraph{Professions}

	We consider the list of professions used in \citet{BCZ16} and \citet{ZZL18}\footnote{\url{https://github.com/tolga-b/debiaswe/tree/master/data/professions.json}} in light of the neighbours-based bias definition.
	Figure~\ref{prof_graphs} plots the professions, with axis X being the original bias and axis Y being the number of male neighbors, before and after debiasing. For both methods, there is a clear correlation between the two variables.
	
	We observe a Pearson correlation of 0.606 (compared to a correlation of 0.747 when checking neighbors according to the biased version) for \textsc{Hard-Debiased} and 0.792 (compared to 0.820) for \textsc{GN-GloVe}. All these correlations are significant with p-values $<1\times10^{-30}$.

	\paragraph{Association between female/male and female/male-stereotyped words}
	We replicate the three gender-related association experiments from \citet{CBN17}. For these experiments we use the full vocabulary since some of the words are not included in the reduced one.
	
	The first experiment evaluates the association between female/male names and family and career words. The second one evaluates the association between female/male concepts and arts and mathematics words. Since the inherently gendered words (e.g. girl, her, brother) in the second experiment are handled well by the debiasing models we opt to use female and male names instead. The third one evaluates the association between female/male concepts and arts and science words. Again, we use female and male names instead.\footnote{All word lists are taken from \citet{CBN17}: \noindent\textbf{First experiment:} \textit{Female names}: Amy, Joan, Lisa, Sarah, Diana, Kate, Ann, Donna. \textit{Male names}: John, Paul, Mike, Kevin, Steve, Greg, Jeff, Bill. \textit{Family words}: home, parents, children, family, cousins, marriage, wedding, relatives. \textit{Career words}: executive, management, professional, corporation, salary, office, business, career. \noindent\textbf{Second experiment:} \textit{Arts Words}: poetry, art, dance, literature, novel, symphony, drama, sculpture. \textit{Math words}: math, algebra, geometry, calculus, equations, computation, numbers, addition. \noindent\textbf{Third experiment:}  \textit{Arts words}: poetry, art, Shakespeare, dance, literature, novel, symphony, drama. \textit{Science words}: science, technology, physics, chemistry, Einstein, NASA, experiment, astronomy.}
	
	For the \textsc{Hard-Debiased} embedding, we get a p-value of $0$ for the first experiment, $0.00016$ for the second one, and $0.0467$ for the third. For the \textsc{GN-GloVe} embedding, we get p-values of $7.7\times10^{-5}$, $0.00031$ and $0.0064$ for the first, second and third experiments, respectively.

	\paragraph{Classifying previously female- and male-biased words}
	Can a classifier learn to generalize from some gendered words to others based only on their representations? 
	We consider the 5,000 most biased words according to the original bias (2,500 from each gender), train an RBF-kernel SVM classifier on a random sample of 1,000 of them (500 from each gender) to predict the gender, and evaluate its generalization on the remaining 4,000. For the \textsc{Hard-Debiased} embedding, we get an accuracy of $88.88\%$, compared to an accuracy of $98.25\%$ with the non-debiased version. For the \textsc{GN-GloVe} embedding, we get an accuracy of $96.53\%$, compared to an accuracy of $98.65\%$ with the non-debiased version.
	
	\section{Discussion and Conclusion}
	\label{discussion}
	
	The experiments described in the previous section reveal a systematic bias found in the embeddings, which is independent of the gender direction. We observe that semantically related words still maintain gender bias both in their similarities, and in their representation. Concretely, we find that: 
	
	\begin{enumerate}
		\item Words with strong previous gender bias (with the same direction) are easy to cluster together.
		\item Words that receive implicit gender from social stereotypes (e.g. receptionist, hairdresser, captain) still tend to group with other implicit-gender words of the same gender, similar as for non-debiased word embeddings.
		\item The implicit gender of words with prevalent previous bias is easy to predict based on their vectors alone.
	\end{enumerate}
	
	The implications are alarming: while suggested debiasing methods work well 
	at removing the gender direction, the debiasing is mostly superficial. The bias stemming from world stereotypes and learned from the corpus is ingrained much more deeply in the embeddings space.
	
	We note that the real concern from biased representations is not the association of a concept with words such as ``he", ``she", ``boy", ``girl" nor being able to perform gender-stereotypical word analogies. While these are nice ``party tricks", algorithmic discrimination is more likely to happen by associating one implicitly gendered term with other implicitly gendered terms, or picking up on gender-specific regularities in the corpus by learning to condition on gender-biased words, and generalizing to other gender-biased words (i.e., a resume classifier that will learn to favor male over female candidates based on stereotypical cues in an existing---and biased---resume dataset, despite of being ``oblivious'' to gender). Our experiments show that such classifiers would have ample opportunities to pick up on such cues also after debiasing w.r.t the gender-direction.
	
	The crux of the issue is that the gender-direction provides a way to \emph{measure} the gender-association of a word, but \emph{does not determine} it.
	Debiasing methods which directly target the gender-direction are for the most part merely hiding the gender bias and not removing it. The popular definitions used for quantifying and removing bias are insufficient, and other aspects of the bias should be taken into consideration as well.	
	
	\section*{Acknowledgments}
	
	This work is supported by the Israeli Science Foundation (grant number 1555/15), and by the Israeli ministry of Science, Technology and Space through the Israeli-French Maimonide Cooperation program.

	\bibliography{bib_full}
	\bibliographystyle{acl_natbib}

\end{document}